\documentclass{article}

\PassOptionsToPackage{numbers, compress}{natbib}

\usepackage[final]{neurips_2024}

\usepackage[utf8]{inputenc} 
\usepackage[T1]{fontenc}    
\usepackage{hyperref}       
\usepackage{url}            
\usepackage{booktabs}       
\usepackage{amsfonts}       
\usepackage{nicefrac}       
\usepackage{microtype}      
\usepackage{xcolor}         
\usepackage{algorithm}
\usepackage{algorithmic}
\usepackage{amsmath}
\usepackage{multicol}
\usepackage{natbib}
\usepackage{graphicx}
\usepackage{caption}
\usepackage{listings}
\title{Where Did It All Go Wrong? A Hierarchical Look into Multi-Agent Error Attribution}

\author{
    Adi Banerjee$^*$ \\
    Amazon Web Services \\
    New York, NY, USA \\
    \texttt{adibaner@amazon.com} \And
    Anirudh Nair$^*$ \\
    Amazon Web Services \\
    Boston, MA, USA \\
    \texttt{rianina@amazon.com} \And
    Tarik Borogovac \\
    Amazon Web Services \\
    Boston, MA, USA \\
    \texttt{tarikbo@amazon.com}
}

\begin{document}
\maketitle
\def\thefootnote{*}\footnotetext{Equal Contribution.}\def\thefootnote{\arabic{footnote}}

\begin{abstract}
  Error attribution in Large Language Model (LLM) multi-agent systems presents a significant challenge in debugging and improving collaborative AI systems. Current approaches to pinpointing agent and step level failures in multi-agent interaction traces\textemdash whether using all-at-once evaluation, step-by-step analysis, or binary search\textemdash fall short when analyzing complex patterns, struggling with both accuracy and consistency. We present ECHO (Error attribution through Contextual Hierarchy and Objective consensus analysis), a novel algorithm that combines hierarchical context representation, objective analysis-based evaluation, and consensus voting to improve error attribution accuracy. Our approach leverages a positional-based leveling of contextual understanding while maintaining objective evaluation criteria, ultimately reaching conclusions through a consensus mechanism. Experimental results demonstrate that ECHO outperforms existing methods across various multi-agent interaction scenarios, showing particular strength in cases involving subtle reasoning errors and complex interdependencies. Our findings suggest that leveraging these concepts of structured, hierarchical context representation combined with consensus-based objective decision-making, provides a more robust framework for error attribution in multi-agent systems.


\end{abstract}


\section{Introduction}


The evolution of large language models (LLMs) has driven their implementation as collaborative, specialized agents working together in structured systems \cite{guo2024large, li2025review}. These systems break down complex challenges into manageable components, distributing them among purpose-specific agents that work in concert toward common objectives \cite{chen2024internet, li2023camel}. Such coordination demands effective orchestration, with each agent performing specialized functions. Building these collaborative systems requires careful consideration of interconnected graph structures\textemdash mapping agent relationships, information flows, and logical constraints. These structural elements form the foundation for expandable and flexible multi-agent frameworks \cite{zhou2025multi}.


Multi-Agent Systems (MASs) have demonstrated remarkable performance across use-cases such as coding \cite{hong2023metagpt}, medical QA \cite{kim2024mdagents}, and financial decision-making \cite{yu2024fincon}. However, their multi-step nature makes them vulnerable to compounding errors, where early mistakes amplify through subsequent steps and can derail the entire system. As a result, identifying the initial error's source—both agent and step—becomes crucial for mitigating such failures and improving these systems.


As MASs grow in complexity, manual error attribution becomes unscalable, necessitating an automated approach. However, according to the Who\&When benchmark \cite{zhang2025agent}, even SOTA LLMs\textemdash both closed-source (GPT-4o \cite{hurst2024gpt}, o1 \cite{jaech2024openai}) and open-source (Llama 4 \cite{meta2025llama})\textemdash struggle with this task. The complexity stems from interdependent agent interactions, large context sizes, and the need to understand both local and global context within interaction traces. Traditional debugging approaches falter in these dynamic systems where errors are often subtle and context-dependent.



Automated error attribution in LLM-based MAS has explored varying approaches to failure log analysis. For example, all-at-once methods expose LLMs to complete logs simultaneously for agent and step identification \cite{zhang2025agent}. Alternatively, step-by-step approaches evaluate interactions sequentially until detecting an error \cite{zhang2025agent}. More sophisticated binary search methods iteratively narrow the search space by having LLMs determine which half of the trace contains the critical mistake \cite{zhang2025agent}.

This paper presents ECHO (Error attribution through Contextual Hierarchy and Objective consensus analysis), a novel approach to error attribution in multi-agent systems, that addresses these limitations, by guiding error attribution through developing a hierarchical context representation of the entire interaction trace, providing independent objective analyses across these contexts and cross-validating their findings via consensus voting.


\section{Related Work}

\subsection{Evaluation of LLM Agents}
Much research has focused on automated LLM evaluation across various tasks \cite{fu2023gptscore,liu2023g,li2023alpacaeval}, with recent efforts shifting toward assessing the agentic abilities of LLMs. AgentBench \cite{liu2023agentbench} tests task completion across environments like online shopping and householding, while MLAgentBench \cite{huang2023mlagentbench} evaluates LLM agents' abilities to perform machine learning experimentation. Similarly, AssistantBench \cite{yoran2024assistantbench} features realistic, time-intensive web tasks requiring multi-turn interactions, while Agent-as-a-Judge \cite{zhuge2024agent} leverages LLM agents to evaluate other agent peers through tool interactions. Finally, ToolFormer \cite{schick2023toolformer}, AnyTool \cite{du2024anytool}, and ToolBench \cite{qin2023toolllm} benchmark LLMs' API-calling capabilities via tool-use. As LLM agent interactions evolve, evaluation frameworks now address multi-agent scenarios: MultiAgentBench \cite{zhu2025multiagentbench} measures coordination and competitive dynamics, while SwarmBench \cite{ruan2025benchmarking} assesses swarm intelligence through coordination tasks like foraging and flocking.

\subsection{Error Attribution}
While LLM evaluation typically measures task completion and output quality, error attribution focuses on systematically classifying error types and patterns to understand fundamental failure modes \cite{xu2025diagnosing, yin2022seq2seq, jiang2023tigerscore, ladhak2022contrastive}. ReaLMistake \cite{kamoi2024evaluating} assesses LLMs' error attribution capabilities in generated output, while SynCheck \cite{wu2024synchronous} leverages decoding dynamics to verify sentence trustworthiness and backtracks the unfaithfulness of generated sentences. Self-Backtracking \cite{yang2025step} trains LLMs to localize and correct their own reasoning errors through supervised fine-tuning. Furthermore, Process Reward Models (PRMs) \cite{lightman2023let} evaluate intermediate reasoning step correctness through process annotation, and ProcessBench \cite{zheng2024processbench} measures PRMs' step-wise evaluation capabilities. These approaches focus solely on single-agent error analysis, whereas the Who\&When dataset \cite{zhang2025agent} and ECHO extends error attribution to multi-agent settings by identifying both the responsible agent and erroneous step.

\section{ECHO Methodology}

Error attribution in multi-agent systems demands three fundamental capabilities: context understanding to capture interaction patterns, error analysis to detect failure points, and decision synthesis to determine final attribution. ECHO addresses these through hierarchical context representation, decoupled objective analysis (at both agent and step levels), and confidence-weighted consensus voting, each targeting specific attribution challenges.
\begin{figure}[htbp]
    \centering
    \captionsetup{labelformat=empty}
    \includegraphics[width=\linewidth]{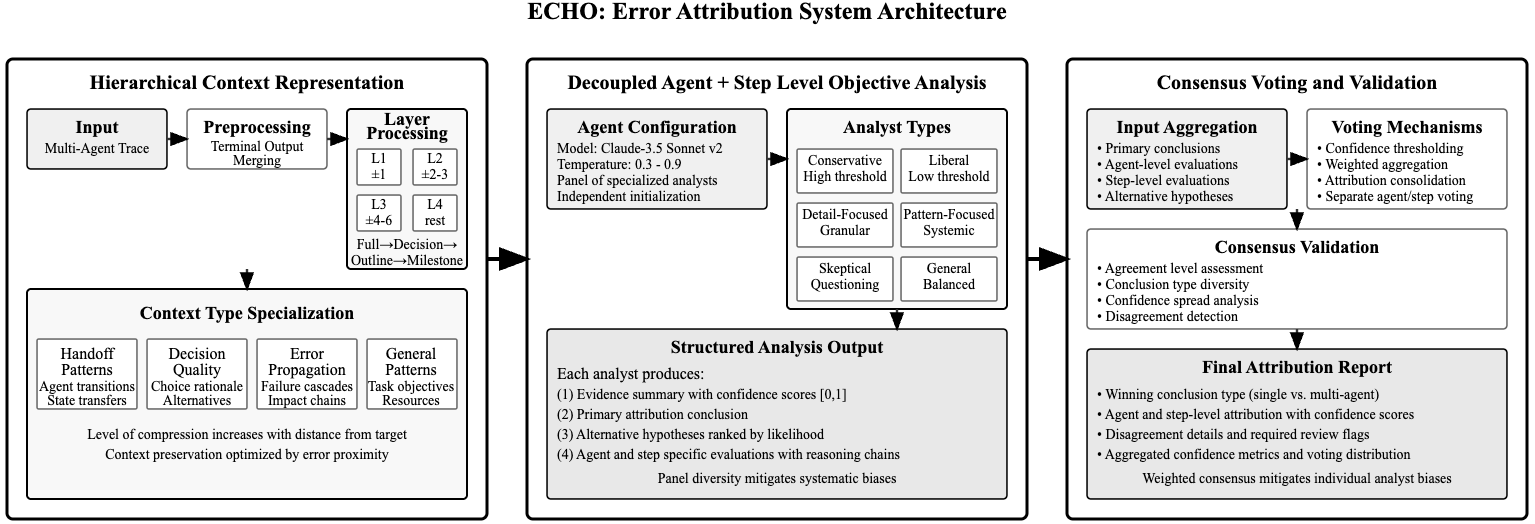}
    \caption{Figure 1: ECHO Architecture. The system comprises: (1) Hierarchical Context - processes traces through 4 compression layers (L1-L4: full content → milestones) with specialized modules for handoffs, decisions, errors, and patterns; (2) Decoupled Analysis - uses 6 specialized agents (conservative to balanced) generating structured outputs with evidence, confidence scores, and hypotheses; (3) Consensus Voting - aggregates analyses via confidence-weighted voting and disagreement resolution.}
    \label{fig:figure_1}
\end{figure}
\subsection{Hierarchical Context Representation}

Error attribution in multi-agent systems faces a fundamental tension: while comprehensive context is crucial for accurate attribution, processing limitations make it impractical to analyze full interaction traces $\tau$ of $n$ agents with equal detail. Traditional approaches typically restrict analysis to immediate positional neighbors (±1 agent), presenting critical limitations. This narrow window fails to capture long-range dependencies where errors propagate across multiple steps and misses crucial context from causally relevant agents outside this window (e.g. error in initial problem formulation manifesting several steps later, or early incorrect assumptions propagating through multiple agents before causing a failure). ECHO addresses these limitations through a multi-layered hierarchical context representation ($C$) that captures both local and global interaction patterns.



Hierarchical representation operates at 4 levels $L_1$ through $L_4$, as seen in Appendix \ref{appendix:echo_algorithm} and \ref{appendix:hierarchical_context}, to extract key information from agent / step interactions (via regex pattern matching). While LLMs could potentially be used for more flexible content extraction, the computational overhead would be significant\textemdash a trade-off we explore further in Section 4.3. Its implementation employs specialized content extraction mechanisms for each layer $C_i$, as seen below:

\begin{enumerate}
    \item \textbf{Immediate Context layer ($L_1$)}: Encompasses the target agent $i$ and its direct neighbors ($\tau_{i\pm1}$), preserving complete reasoning chains and interaction patterns. This layer maintains full agent content, including detailed decision-making processes, input-output relationships, and intermediate computations. This granular preservation enables precise analysis of local decision points and immediate error manifestations. Examples of these are: "Let me analyze the data... [Terminal Output]: DataFrame loaded, showing 500 rows". 
    \item \textbf{Local Context Layer ($L_2$)}: Spans $\tau_{i\pm2,3}$ steps from the target agent, focusing on tactical decision sequences and their interconnections. By preserving key reasoning chains while filtering routine operations, this layer reveals short-range error propagation patterns and local decision dependencies. This intermediate scope bridges the gap between immediate interactions and broader strategic patterns. Examples of these are focusing on specific patterns such as conclusive statements (e.g., "I conclude the optimal parameters are x=0.5, y=1.0") and logical transitions (e.g., "Therefore, we should use gradient descent"). 
    \item \textbf{Distant Context Layer ($L_3$)}: Covers $\tau_{i\pm4,5,6}$ steps through strategic compression, distilling agent interactions into concise outcome summaries. This layer captures essential state transitions, critical assumptions, and warning signals while filtering peripheral details. This compression enables identification of longer-range dependencies without overwhelming the analysis with excessive detail. Examples of these are critical state changes (e.g., "Model training failed: insufficient data"), error conditions (e.g., "Classification blocked: missing features"), and handoff information (e.g., "Received dataset with normalized features"). 
    \item \textbf{Global Context Layer ($L_4$)}: Encompasses the remaining interaction trace ($\tau_{remainder}$) through highly compressed milestone-based representation. By retaining only strategically significant decision points and major state transitions, this layer enables system-wide consistency checking and identification of broad error patterns. This high-level view ensures local decisions align with global objectives while maintaining computational feasibility. Examples of these include error propagation signals (e.g., "Persistent data quality errors"), major state transitions (e.g., "Switched from classification to regression"), and cross-agent dependencies (e.g., "Reusing parameters from Agent 2's optimization"). 
\end{enumerate}

This graduated extraction and compression approach adapts to different context types (handoff, decision quality, error propagation, general), enabling context-aware information preservation at each layer. Lastly, it is worth noting that when using regex-based extraction, real-world systems can be designed to generate reasoning traces with extractable keywords, effectively pre-seeding content for regex. This maintains efficiency while achieving accuracy comparable to more complex methods, making it practical for real applications.


\subsection{Objective Analysis}


ECHO employs a panel of $k$ objective analysis agents that evaluate the full interaction trace $\tau$ through hierarchical context $C$ (algorithm in Appendix \ref{appendix:echo_algorithm}). Each analysis agent independently assesses all steps and provides confidence scores $\sigma_j$, enabling nuanced attributions that can identify distributed responsibility, interaction effects, and systemic issues spanning multiple agents or steps.

The objective analysis framework leverages the hierarchical context through conversation summaries that preserve contextual relationships, enabling multi-scale error attribution. Agents examine errors across all layers: detailed decisions ($L_1$), state transitions ($L_2$), error propagation ($L_3$), and systemic issues ($L_4$). Furthermore, attribution is restricted to steps with explicit reasoning, ensuring precise analysis across context scales while maintaining step-level granularity.

The effectiveness of objective analysis depends critically on agent diversity to mitigate systematic biases. Rather than deploying identical agents that might propagate the same analytical biases, ECHO employs a panel of specialized analysts with roles $\rho_j$ (as seen in the code in Appendix \ref{appendix:objective_analysis}):
\begin{enumerate}
    \item \textbf{Conservative Analyst}: Requires strong, clear evidence for attribution, prefers single-agent attributions, maintains high confidence thresholds, and only attributes with definitive proof
    \item \textbf{Liberal Analyst}: Considers multi-agent error scenarios, identifies subtle error patterns, accepts moderate confidence thresholds, and makes attributions with reasonable evidence
    \item \textbf{Detail-Focused Analyst}: Examines specific evidence and exact wording, identifies subtle inconsistencies, focuses on fine-grained analysis, and prioritizes concrete evidence over patterns
    \item \textbf{Pattern-Focused Analyst}: Recognizes broader reasoning chains, tracks error propagation patterns, identifies recurring themes, and analyzes overall reasoning structure
    \item \textbf{Skeptical Analyst}: Questions underlying assumptions, explores alternative explanations, challenges conventional attributions, and examines validity of ground truth
    \item \textbf{General Analyst}: Maintains balanced perspective, considers all evidence types equally, focuses on obvious, impactful errors, and provides baseline objective evaluation
\end{enumerate}

Each analysis agent follows a structured evaluation protocol, examining both individual steps and their hierarchical context relationships. The agents' analyses ($\epsilon_j$) are guided by specialized prompts that reflect their assigned analytical focus, producing outputs with four components: (1) investigation summary, (2) detailed evaluations with error likelihood scores [0,1] and evidence, (3) primary conclusion specifying attribution type, responsible agent(s), mistake step, and confidence score $\sigma_j$, and (4) alternative hypotheses. This structure ensures consistent, comparable outputs for consensus voting across diverse analytical perspectives.

This approach prevents echo-chamber effects in attribution through intentional perspective diversity. While this was implemented in ECHO through role-specific prompts and temperatures, the framework accommodates other diversification methods (different models, LLM configurations, or reasoning strategies)\textemdash all serving the core principle of bias mitigation through analytical diversity.

\subsection{Consensus Voting}
The final component of ECHO is a consensus voting mechanism that aggregates analyses from the panel of $k$ objective analysts, as seen in Appendix \ref{appendix:consensus_voting}. The voting system operates through weighted confidence consensus, where each analyst's attribution is weighted by their reported confidence level $\sigma_j$, subject to a minimum confidence threshold $\delta$ to filter out low-confidence attributions.

The consensus mechanism processes 3 key components from each analysis $A_t^j$ (where $t \in \{\text{agent}, \text{step}\}$): (1) Primary Conclusions\textemdash that considers core attribution decisions including the attribution type, responsible agent(s), and specific error step; (2) Agent Evaluations\textemdash that provides assessments of each agent's error likelihood and supporting evidence; and (3) Alternative Hypotheses\textemdash that considers secondary attribution possibilities that maintain reasonable likelihood.

The voting process follows a hierarchical decision structure, that first determines the winning conclusion type through weighted confidence aggregation of each analyzer's agent and step vote ($V_a, V_s$). Then for single/multi-agent conclusions, ECHO identifies specific agents through confidence-weighted attribution votes ($\omega_a$), after which it validates and aggregates step-level predictions ($\omega_s$). Finally it synthesizes supporting reasoning from the highest-confidence attributions.

The system explicitly handles disagreements through disagreement analysis $\phi$ by examining: (1) conclusion diversity (number of different conclusion types); (2) confidence spread across analysts ($\max(\sigma_j) - \min(\sigma_j)$); (3) attribution consistency between agent and step-level predictions; (4) need for additional review when high disagreement exists (spread > 0.5) or multiple conflicting high-confidence attributions emerge.

This structured voting approach ensures that the final attribution (ConsensusResult($\omega_a, \omega_s, \phi$)) reflects the collective analysis while maintaining sensitivity to strong minority opinions and potential edge cases. The combination of confidence weighting and explicit disagreement handling provides robustness against individual analyst biases while preserving valuable insights from dissenting views.

\section{Results and Analysis}
\subsection{Experimental Setup}

We evaluate our implementations using the Who\&When benchmark \cite{zhang2025agent}, a comprehensive dataset comprising failure logs from LLM-powered multi-agent systems. This benchmark is particularly suitable for error attribution evaluation as it provides fine-grained failure annotations across diverse scenarios, including both algorithm-generated and hand-crafted multi-agent interactions.

\textbf{Dataset Characteristics} - The benchmark consists of 2 distinct subsets: (1) Algorithm-Generated: systematic failure logs from automated multi-agent interactions; (2) Hand-Crafted: curated failure scenarios designed to capture complex error patterns. Each failure log is annotated with the failure-responsible agent; the specific step where the error occurred; and detailed failure reasoning.

\textbf{Evaluation Protocol} - We assess each implementation across 4 experimental conditions: (1) Algorithm-Generated with ground truth; (2) Algorithm-Generated without ground truth; (3) Hand-Crafted with ground truth; and (4) Hand-Crafted without ground truth. Attribution accuracy is measured at 2 levels: agent-level (referring to the correct identification of the failure-responsible agent; and step-level (referring to the correct identification of the specific error step).



\textbf{Implementation Details} - Our framework employs Claude 3.5 Sonnet v2 as the base LLM, with varying temperatures between 0.3 and 0.9 across 3 randomly sampled analysis agents (from a pool of 6 specialists) for every scenario, to balance analytical diversity while reducing computational overhead. We use a confidence threshold ($\delta = 0.3$) for filtering attributions in the consensus voting mechanism, to balance inclusion of diverse perspectives while filtering out highly uncertain predictions. This implementation assumes position-based contextual importance, independence between analysis agents, and treats errors as primarily additive rather than emergent.

\subsection{Comparative Analysis of Implementations}

We evaluate 4 progressive implementations of error attribution, each building upon the limitations of its predecessor:

\textbf{Implementation 1 (I1) - Fixed Context Window}:
Represents the baseline approach using a fixed context window (±1 step), where a context-aware agent analyzes each step with its immediate neighbors, followed by a final judge agent for attribution decisions.

\textbf{Implementation 2 (I2) - Hierarchical Context}:
Enhances I1 by replacing the fixed context window with a four-layer hierarchical context representation, maintaining the same attribution mechanism but providing graduated access to the full interaction trace.

\textbf{Implementation 3 (I3) - Objective Analysis}:
Builds upon I2's hierarchical context by replacing the context-aware agent and judge agent with a panel of specialized objective analyst agents and performing consensus voting on those outcomes, enabling diverse analytical perspectives.

\textbf{Implementation 4 (I4) - Decoupled Attribution}:
Refines I3 by separating the attribution process into two distinct phases: agent-level attribution to identify responsible agents, followed by step-level attribution to pinpoint specific error points, allowing for more focused analysis at each level.

\subsubsection{Performance of ECHO}

\begin{table}[t]
\caption{Performance of ECHO across different datasets and configurations}
\label{tab:echo-results}
\centering
\small
\begin{tabular}{lccccc}
\toprule
& \multicolumn{2}{c}{Hand-Crafted Dataset} & \multicolumn{2}{c}{Algorithm-Generated Dataset} & \\
\cmidrule(lr){2-3} \cmidrule(lr){4-5}
Method & With GT & Without GT & With GT & Without GT & P-value$^{\dagger}$ \\
\midrule
\multicolumn{6}{l}{\textbf{Agent-Level Accuracy}} \\
Random & 0.120 & 0.120 & 0.291 & 0.291 & <0.001 \\
All-at-Once & 0.577 & 0.529 & 0.563 & 0.530 & 0.032 \\
Step-by-Step & 0.360 & 0.343 & 0.397 & 0.283 & <0.001 \\
Binary Search & 0.517 & 0.362 & 0.441 & 0.301 & 0.007 \\
\midrule[\heavyrulewidth]
\textbf{ECHO (ours)} & \textbf{0.684} & \textbf{0.679} & \textbf{0.688} & \textbf{0.672} & - \\
\midrule
\multicolumn{6}{l}{\textbf{Step-Level Accuracy (Exact)}} \\
Random & 0.042 & 0.042 & 0.191 & 0.191 & <0.001 \\
All-at-Once & 0.060 & 0.021 & 0.152 & 0.145 & <0.001 \\
Step-by-Step & 0.066 & 0.069 & 0.274 & 0.178 & 0.003 \\
Binary Search & 0.069 & 0.069 & 0.240 & 0.166 & 0.012 \\
\midrule[\heavyrulewidth]
\textbf{ECHO (ours)} & \textbf{0.281} & \textbf{0.268} & \textbf{0.288} & \textbf{0.272} & - \\
\midrule
\multicolumn{6}{l}{\textbf{Step-Level with Tolerance} \textit{(Hand-Crafted with GT)}} \\
& All-at-Once & Step-by-Step & Binary Search & \textbf{ECHO (ours)} & P-value$^{\dagger}$ \\
±1 step & 0.149 & 0.166 & 0.138 & \textbf{0.351} & <0.001 \\
±2 steps & 0.223 & 0.189 & 0.190 & \textbf{0.386} & 0.004 \\
±3 steps & 0.350 & 0.209 & 0.224 & \textbf{0.421} & 0.029 \\
±4 steps & 0.380 & 0.294 & 0.319 & \textbf{0.579} & 0.006 \\
±5 steps & 0.428 & 0.351 & 0.362 & \textbf{0.614} & 0.008 \\
\midrule
\multicolumn{6}{l}{\textbf{Token Cost} \textit{(Hand-Crafted with GT)}} \\
& All-at-Once & Step-by-Step & Binary Search & \textbf{ECHO (ours)} & - \\
Tokens & \textbf{17,106} & 87,720 & 34,659 & 53,701 & - \\
\bottomrule
\multicolumn{6}{l}{$^{\dagger}$P-values compare ECHO against each baseline using chi-squared test.} \\
\end{tabular}
\end{table}

ECHO demonstrates robust statistically significant performance (P < 0.05) across both hand-crafted and algorithm-generated datasets, with several notable patterns emerging from the results:

\textbf{Agent-Level Attribution}:

The system achieves consistent agent-level accuracy ($\sim$68\%) across all configurations, with minimal degradation (1--2\%) when ground truth is withheld. Furthermore, the similar performance across both dataset types (hand-crafted: 68.4\%, algorithm-generated: 68.8\%) further demonstrates ECHO's robust generalization capabilities across different interaction patterns. This ground truth independence, combined with consistent cross-dataset performance, makes ECHO particularly valuable for deployment in dynamic, real-world AI systems where there is generally an absence of labeled data.

\textbf{Step-Level Attribution}:
Although ECHO still outperforms prior methods with statistical significance, step-level precision still proves challenging, with exact matches at 27--28\%. However, accuracy improves significantly when accounting for step tolerance. With ±3 steps, accuracy reaches 42.1\% for hand-crafted dataset with ground truth; and when further extended to ±5 steps, accuracy reaches 61.4\%\textemdash again, maintaining superior performance over baselines.

\textbf{Impact of Ground Truth}:
The minimal difference in performance between configurations with and without ground truth (typically <2\% variation) suggests that ECHO's architecture effectively leverages interaction patterns and context for attribution, rather than relying on ground truth information. This is particularly important for real-world applications where ground truth may not be available.

\textbf{Token Counts}:
ECHO demonstrates moderate token efficiency, using $\sim$54K tokens compared to $\sim$88K for Step-by-Step and $\sim$17K for All-at-Once approaches on hand-crafted datasets. While higher than the simplest baseline, ECHO's token usage reflects its balanced approach to comprehensive analysis, maintaining reasonable processing costs of \$0.15 per example with Claude 3.5 Sonnet v2.

The results demonstrate ECHO's capability to provide reliable agent-level attribution while offering flexible step-level identification with tolerance ranges. The system's consistent performance across different datasets and configurations, combined with its reasonable computational overhead, makes it suitable for practical applications in error attribution for multi-agent systems.

\subsection{Impact of ECHO Components Via Ablation}


\begin{table}[t]
\caption{Ablation Study: Impact of Each Component}
\label{tab:ablation-combined}
\centering
\small
\begin{tabular}{lccccc}
\toprule
& \multicolumn{2}{c}{Hand-Crafted Dataset} & \multicolumn{2}{c}{Algorithm-Generated Dataset} & \\
\cmidrule(lr){2-3} \cmidrule(lr){4-5}
Implementation & With GT & Without GT & With GT & Without GT & P-value$^{\ddagger}$ \\
\midrule
\multicolumn{6}{l}{\textbf{Agent-Level Accuracy}} \\
Fixed Context (I1)$^{\dagger}$ & 0.286 & 0.265 & 0.461 & 0.452 & - \\
+ Hierarchical (I2)$^{\dagger}$ & 0.447 & 0.429 & 0.523 & 0.508 & 0.037 \\
+ Objective Analysis (I3) & 0.610 & 0.589 & 0.651 & 0.635 & 0.043 \\
+ Decoupled Attribution (I4) & \textbf{0.684} & \textbf{0.679} & \textbf{0.688} & \textbf{0.672} & 0.196 \\
\midrule
\multicolumn{6}{l}{\textbf{Step-Level Accuracy}} \\
Fixed Context (I1)$^{\dagger}$ & 0.151 & 0.143 & 0.157 & 0.140 & - \\
+ Hierarchical (I2)$^{\dagger}$ & 0.170 & 0.166 & 0.192 & 0.175 & 0.398 \\
+ Objective Analysis (I3) & 0.232 & 0.218 & \textbf{0.461} & \textbf{0.444} & <0.001 \\
+ Decoupled Attribution (I4) & \textbf{0.281} & \textbf{0.268} & 0.288 & 0.272 & 0.211 \\
\midrule
\multicolumn{6}{l}{\textbf{Token Cost}} \\
Fixed Context (I1)$^{\dagger}$ & 4.02M & 3.93M & 319K & 317K & - \\
+ Hierarchical (I2)$^{\dagger}$ & 7.70M & 7.66M & 407K & 405K & - \\
+ Objective Analysis (I3) & 67.5K & 66.5K & \textbf{12.6K} & \textbf{12.5K} & - \\
+ Decoupled Attribution (I4) & \textbf{33.0K} & \textbf{32.5K} & 12.8K & 12.7K & - \\
\bottomrule
\multicolumn{6}{l}{\small $^{\dagger}$Hand-Crafted Dataset results for I1 and I2 based on limited sample of shorter traces} \\
\multicolumn{6}{l}{\small $^{\ddagger}$P-values compare each component with previous implementation} \\
\end{tabular}
\end{table}

\textbf{The Question of Unifying or Decoupling Objective Analyses}

The analysis of context length reveals key insights about unified versus decoupled attribution (Table \ref{tab:ablation-combined}). For shorter algorithm-generated traces, unified analysis achieves strong results (65.1\% agent-level, 46.1\% step-level accuracy), while decoupling shows mixed effects: marginal improvement in agent-level accuracy (68.8\%) but notable decline in step-level precision (28.8\%). This suggests that splitting attribution tasks can be counterproductive when operating within the LLM's comfortable processing range ($\sim$13K tokens unified vs. $\sim$6K+7K tokens decoupled).


For longer hand-crafted traces, unified analysis struggles with step-level precision (23.2\%) while maintaining moderate agent-level accuracy (61.0\%). Decoupling improves performance (68.4\% agent-level, 28.1\% step-level) while significantly reducing per-task tokens (from $\sim$67K to $\sim$13K+19K). This reveals context-length dependency: shorter traces benefit from unified analysis's complete context, while longer traces near model limits favor decoupled analysis' complexity management.

\textbf{The Computational Overhead Resolution By Using Objective Analysis}

The shift to objective analysis (I3) reveals key limitations of context-aware agents in error attribution. While fixed (I1) and hierarchical (I2) context implementations leverage trace information, their repeated analysis proves computationally intensive. Objective analysis dramatically improves efficiency, reducing token usage by 60-110x for hand-crafted cases, and by 25-30x for algorithm-generated cases, while still improving accuracy ($\sim$+16.3\% agent-level and $\sim$+6.2\% step-level at P < 0.05) as shown in Table \ref{tab:ablation-combined}. These results demonstrate that objective analysis is crucial for practical deployment, enabling efficient processing of longer traces while maintaining comprehensive context benefits.\newline

\textbf{Switching from Fixed-Window Context to Hierarchical Context}

Hierarchical context shows clear advantages over fixed context windows within the context-aware framework. Shifting from fixed ±1 step (I1) to hierarchical context (I2) yields significant improvements in hand-crafted datasets: +16.1\% in agent-level accuracy (P < 0.05) and +1.9\% in step-level accuracy (not significant), as shown in Table \ref{tab:ablation-combined}. A similar trend is exhibited for the algorithm-generated dataset, with a +6.2\% in agent-level accuracy and +3.8\% in step-level accuracy.  This validates graduated context preservation, where detail decreases with distance from target steps. While analysis for the hand-crafted dataset was limited to shorter traces due to computational constraints with context-aware agents, the consistent accuracy improvements highlight hierarchical context's value\textemdash a benefit fully realized when combined with objective analysis.

\begin{table}[t]
\caption{Additional Ablations on Hand-Crafted Dataset with Ground Truth}
\label{tab:ablation-additional}
\centering
\small
\begin{tabular}{lcccc}
\toprule
Implementation & Agent-Level & Step-Level & Token Cost & P-value$^{\ddagger}$ \\
\midrule
\multicolumn{5}{l}{\textbf{LLM Capability}} \\
Claude-3 Haiku & 0.618 & 0.164 & 60,008 & 0.042 \\
\textbf{ECHO (Claude-3.5 Sonnet)} & 0.684 & \textbf{0.281} & \textbf{53,701} & - \\
Claude-3.7 Sonnet & \textbf{0.788} & 0.269 & 57,223 & 0.036 \\
\midrule
\multicolumn{5}{l}{\textbf{Hierarchical Context Method}} \\
\textbf{ECHO (Regex-based)} & 0.684 & 0.281 & \textbf{53,701} & - \\
LLM-based$^{\dagger}$ & \textbf{0.750} & \textbf{0.438} & 165,932 & - \\
\midrule
\multicolumn{5}{l}{\textbf{Objective Analyst Panel Size}} \\
\textbf{ECHO (3 Analysts)} & \textbf{0.684} & \textbf{0.281} & \textbf{53,701} & - \\
6 Analysts & 0.667 & 0.259 & 87,603 & 0.483 \\
\bottomrule
\multicolumn{5}{l}{\small $^{\dagger}$Tested on limited sample of shorter traces due to computational constraints} \\
\multicolumn{5}{l}{\small $^{\ddagger}$P-values compare with ECHO baseline for each sections} \\
\end{tabular}
\end{table}

\textbf{Additional Ablation Studies}

Additional ablation studies on the hand-crafted dataset with ground truth reveal key design insights. Using stronger reasoning models significantly improve performance (P < 0.05): Claude-3.7 Sonnet achieves 78.8\% agent-level and 26.9\% step-level accuracy versus Haiku's 61.8\% and 16.4\%, maintaining similar token usage ($\sim$60K). Switching from regex-based context extraction to LLM-based context extraction shows promise (75.0\% agent-level, 43.8\% step-level) but proves computationally intractable ($\sim$166K vs. 54K tokens, with extraction alone using $\sim$145K). Lastly, expanding the analyst panel size from 3 to 6 yields minimal gains (68.4\% vs. 66.7\% agent-level) while doubling token usage (88K vs. 54K), validating our default configuration's efficiency. These results have been shown in Table \ref{tab:ablation-additional}.

\section{Conclusion}

We introduce ECHO, a novel approach to error attribution in multi-agent systems that combines hierarchical context representation, decoupled objective analysis, and confidence-weighted consensus voting. Our results demonstrate substantial improvements over existing baseline methods, particularly for longer traces where traditional approaches become prohibitive. Beyond immediate applications, ECHO's precise error attribution capabilities have broader implications for AI development: in reinforcement learning, it can help identify and eliminate false steps that may reduce model efficacy, while in single-agent optimization, it enables targeted prompt refinement for systems like GEPA \cite{agrawal2025gepareflectivepromptevolution} and DEEVO \cite{nair2025tournamentpromptsevolvingllm}. Future directions include developing relevance-based architectures for dynamic context preservation, enhancing the consensus mechanism through multi-agent debate protocols \cite{nair2025tournamentpromptsevolvingllm}, and incorporating error severity metrics as well as partial correctness evaluation. Additionally, the Who\&When benchmark could be expanded with standardized problem-type categorization and complex architectural patterns to better reflect real-world scenarios. As multi-agent systems proliferate, ECHO's efficient context handling and bias mitigation approach provides a crucial foundation for both debugging current systems and advancing AI development practices.

\newpage
\bibliographystyle{IEEEtran}
\bibliography{sample-base}





\appendix

\section{Appendix}

\subsection{ECHO Algorithm}\label{appendix:echo_algorithm}

\begin{algorithm}[H]
\caption{ECHO: Error Attribution through Contextual Hierarchy and Objective Consensus Analysis}
\begin{algorithmic}[1]
\vspace{0.5cm}
\REQUIRE
    \STATE $\tau$ : interaction trace of $n$ agents
    \STATE $\alpha$ : final answer
    \STATE $\delta$ : minimum confidence threshold
    \STATE $k$ : number of analysis agents
    \STATE $\gamma$ : ground truth (optional)
\vspace{0.4cm}

\ENSURE Attribution of error to specific agent(s) and step(s)

\vspace{0.5cm}

\STATE \textbf{Procedure} HierarchicalContextExtraction($\tau$):
    \STATE $C \leftarrow \emptyset$ \COMMENT{Init context}
    \FOR{each agent $i \in \{1,...,n\}$}
        \STATE $L_1 \leftarrow$ ExtractFullContext($\tau_{i\pm1}$)
        \STATE $L_2 \leftarrow$ ExtractKeyDecisions($\tau_{i\pm2,3}$)
        \STATE $L_3 \leftarrow$ CompressSummaries($\tau_{i\pm4,5,6}$)
        \STATE $L_4 \leftarrow$ ExtractMilestones($\tau_{remainder}$)
        \STATE $C_i \leftarrow \{L_1, L_2, L_3, L_4\}$
    \ENDFOR
    \RETURN $C$

\vspace{0.5cm}

\STATE \textbf{Procedure} DecoupledAgentAndStepAnalysis($C, \alpha, \gamma$):
    \FOR{type $t \in \{\text{agent}, \text{step}\}$}
        \STATE $A_t \leftarrow \emptyset$ \COMMENT{Init results}
        \FOR{each analyst $j \in \{1,...,k\}$}
            \STATE $\rho_j \leftarrow$ AnalystRole($j$)
            \IF{$\gamma \neq \text{None}$}
                \STATE $\epsilon_j \leftarrow$ Eval($t, C, \rho_j, \gamma$)
            \ELSE
                \STATE $\epsilon_j \leftarrow$ Eval($t, C, \rho_j$)
            \ENDIF
            \STATE $\sigma_j \leftarrow$ ConfidenceScore($\epsilon_j$)
            \STATE $A_t^j \leftarrow \{\epsilon_j, \sigma_j\}$
        \ENDFOR
    \ENDFOR
    \RETURN $(A_{\text{agent}}, A_{\text{step}})$

\vspace{0.5cm}

\STATE \textbf{Procedure} ConsensusVoting($A_a, A_s, \delta$):
    \STATE $V_a, V_s \leftarrow \emptyset, \emptyset$ \COMMENT{Init voting}
    \FOR{each analysis pair $(A_a^j, A_s^j)$}
        \IF{$\sigma_j \geq \delta$}
            \STATE $V_a, V_s \leftarrow V_a \cup \{A_a^j\}, V_s \cup \{A_s^j\}$
        \ENDIF
    \ENDFOR
    \STATE $\omega_a, \omega_s \leftarrow$ WeightedAggregate($V_a, V_s$)
    \STATE $\phi \leftarrow$ DisagreementAnalysis($V_a, V_s$)
    \RETURN ConsensusResult($\omega_a, \omega_s, \phi$)

\vspace{0.5cm}

\STATE $C \leftarrow$ HierarchicalContextRepresentation($\tau$)
\STATE $A_a, A_s \leftarrow$ DecoupledAgentAndStepAnalysis($C, \alpha, \gamma$)
\RETURN ConsensusVoting($A_a, A_s, \delta$)
\end{algorithmic}
\end{algorithm}

\subsection{Fixed-Window Context}\label{appendix:fixed_window_context}
\begin{lstlisting}[language=Python, 
                   basicstyle=\ttfamily\footnotesize,
                   breaklines=true,
                   columns=flexible,
                   backgroundcolor=\color{gray!5}
                   ]

def extract_agent_contexts(
    conversation_history: List[Dict[str, Any]]
) -> List[Tuple[Optional[Dict[str, Any]], Dict[str, Any], Optional[Dict[str, Any]]]]:
    """
    Extract agent contexts from conversation history.
    Each context includes the previous agent, current agent, and next agent.

    Args:
        conversation_history: List of conversation turns with agent information

    Returns:
        List of tuples containing (prev_agent, current_agent, next_agent) for each agent
    """

    contexts = []
    for i in range(len(conversation_history)):
        # Get previous agent (None if first agent)
        prev_agent = conversation_history[i - 1] if i > 0 else None

        # Get current agent
        current_agent = conversation_history[i]

        # Get next agent (None if last agent)
        next_agent = conversation_history[i + 1] if i < len(conversation_history) - 1 else None

        contexts.append((prev_agent, current_agent, next_agent))

    return contexts
\end{lstlisting}

\subsection{Hierarchical Context Extraction}\label{appendix:hierarchical_context}
\begin{lstlisting}[language=Python, 
                   basicstyle=\ttfamily\footnotesize,
                   breaklines=true,
                   columns=flexible,
                   backgroundcolor=\color{gray!5}
                   ]
def extract_key_decision(
    agent_content: str, max_words: int = 50, context_type: str = "decision_quality"
) -> str:
    """
    Extract key decision or main point from agent content using regex patterns.

    Args:
        agent_content: The full content of the agent
        max_words: Maximum words in the extracted key decision
        context_type: Type of context to focus on (handoff, decision_quality, error_propagation, general)

    Returns:
        Key decision or main point from the agent's content
    """
    if not agent_content.strip():
        return "No content available"

    if context_type == "handoff":
        patterns = [
            r"(?:received|got|obtained|from)\s+([^.!?]*[.!?])",
            r"(?:passing|providing|sending|to)\s+([^.!?]*[.!?])",
            r"(?:based on|using)\s+([^.!?]*[.!?])",
            r"(?:will|need to|should)\s+([^.!?]*(?:next|continue)[^.!?]*[.!?])",
        ]
    elif context_type == "decision_quality":
        patterns = [
            r"(?:I (?:conclude|determine|decide|believe|think))\s+([^.!?]*[.!?])",
            r"(?:Therefore|Thus|So|Hence),?\s+([^.!?]*[.!?])",
            r"(?:The (?:answer|solution|result))\s+(?:is|appears)\s+([^.!?]*[.!?])",
            r"(?:Based on|Given)\s+([^.!?]*[.!?])",
        ]
    elif context_type == "error_propagation":
        patterns = [
            r"(?:error|mistake|wrong|incorrect|failed)\s+([^.!?]*[.!?])",
            r"(?:cannot|unable|couldn't|can't)\s+([^.!?]*[.!?])",
            r"(?:However|But|Unfortunately)\s+([^.!?]*[.!?])",
        ]
    else:  # general
        patterns = [
            r"(?:I (?:will|should|need to|decided to|conclude that|believe|think|determine)) ([^.!?]*[.!?])",
            r"(?:Therefore|Thus|So|Hence),? ([^.!?]*[.!?])",
            r"(?:The answer|The result|The solution) (?:is|appears to be|seems to be) ([^.!?]*[.!?])",
            r"Let me ([^.!?]*[.!?])",
            r"(?:My approach|My strategy|My plan) (?:is|will be) ([^.!?]*[.!?])",
        ]

    # Try to find pattern matches
    for pattern in patterns:
        matches = re.findall(pattern, agent_content, re.IGNORECASE)
        if matches:
            decision = matches[0].strip()
            words = decision.split()[:max_words]
            return " ".join(words) + ("..." if len(decision.split()) > max_words else "")

    # Fallback: take the first sentence or first max_words
    sentences = agent_content.split(". ")
    if sentences:
        first_sentence = sentences[0].strip()
        if not first_sentence.endswith("."):
            first_sentence += "."
        words = first_sentence.split()[:max_words]
        return " ".join(words) + ("..." if len(first_sentence.split()) > max_words else "")

    # Final fallback: just truncate
    words = agent_content.split()[:max_words]
    return " ".join(words) + ("..." if len(agent_content.split()) > max_words else "")


def summarize_agent(agent_content: str, max_words: int = 20, context_type: str = "general") -> str:
    """
    Create a brief summary of agent content using regex patterns.

    Args:
        agent_content: The full content of the agent
        max_words: Maximum words in the summary
        context_type: Type of context to focus on (handoff, decision_quality, error_propagation, general)

    Returns:
        Brief summary of the agent's content
    """
    if not agent_content.strip():
        return "No content available"

    # Remove excessive whitespace and newlines
    cleaned_content = " ".join(agent_content.split())

    if context_type == "handoff":
        patterns = [
            r"(?:received|got|obtained)\s+([^.!?]*[.!?])",
            r"(?:providing|sending)\s+([^.!?]*[.!?])",
        ]
    elif context_type == "decision_quality":
        patterns = [
            r"(?:conclude|determine|decide)\s+([^.!?]*[.!?])",
            r"(?:Therefore|Thus|So),?\s+([^.!?]*[.!?])",
        ]
    elif context_type == "error_propagation":
        patterns = [
            r"(?:error|mistake|failed)\s+([^.!?]*[.!?])",
            r"(?:cannot|unable)\s+([^.!?]*[.!?])",
        ]
    else:  # general
        patterns = [
            r"(?:In conclusion|To conclude|Therefore|Thus|So|Hence),? ([^.!?]*[.!?])",
            r"(?:The (?:answer|result|solution|output)) (?:is|appears to be|seems to be) ([^.!?]*[.!?])",
            r"(?:I (?:found|determined|concluded|calculated)) ([^.!?]*[.!?])",
        ]

    # Try pattern matching first
    for pattern in patterns:
        matches = re.findall(pattern, cleaned_content, re.IGNORECASE)
        if matches:
            summary = matches[0].strip()
            words = summary.split()[:max_words]
            return " ".join(words) + ("..." if len(summary.split()) > max_words else "")

    # Fallback: take first sentence and truncate
    sentences = cleaned_content.split(". ")
    if sentences:
        first_sentence = sentences[0].strip()
        words = first_sentence.split()[:max_words]
        return " ".join(words) + ("..." if len(first_sentence.split()) > max_words else "")

    # Final fallback
    words = cleaned_content.split()[:max_words]
    return " ".join(words) + ("..." if len(cleaned_content.split()) > max_words else "")


def obtain_milestones(agent_content: str, max_words: int = 15, context_type: str = "general") -> str:
    """
    Extract milestone-based information from agent content using regex patterns.
    This provides a higher level of abstraction than brief summaries for distant contexts.

    Args:
        agent_content: The full content of the agent
        max_words: Maximum words in the extracted milestones
        context_type: Type of context to focus on (handoff, decision_quality, error_propagation, general)

    Returns:
        Milestone-based information from the agent's content
    """
    if not agent_content.strip():
        return "No milestones available"

    # Remove excessive whitespace and newlines
    cleaned_content = " ".join(agent_content.split())

    if context_type == "handoff":
        patterns = [
            r"(?:received|obtained|got)\s+([^.!?]*(?:from|data|information)[^.!?]*[.!?])",
            r"(?:provided|sent|passed)\s+([^.!?]*(?:to|data|information)[^.!?]*[.!?])",
            r"(?:completed|finished)\s+([^.!?]*(?:handoff|transfer)[^.!?]*[.!?])",
        ]
    elif context_type == "decision_quality":
        patterns = [
            r"(?:decided|determined|concluded)\s+([^.!?]*[.!?])",
            r"(?:evaluated|assessed|analyzed)\s+([^.!?]*[.!?])",
            r"(?:final decision|ultimate choice)\s*[:-]?\s*([^.!?]*[.!?])",
        ]
    elif context_type == "error_propagation":
        patterns = [
            r"(?:error|mistake|failure)\s+(?:occurred|detected)\s+([^.!?]*[.!?])",
            r"(?:identified|found)\s+(?:error|issue|problem)\s+([^.!?]*[.!?])",
            r"(?:corrected|fixed|resolved)\s+([^.!?]*[.!?])",
        ]
    else:  # general
        patterns = [
            r"(?:completed|finished|achieved|accomplished)\s+([^.!?]*[.!?])",
            r"(?:created|generated|produced|built)\s+([^.!?]*[.!?])",
            r"(?:step\s+\d+|phase\s+\d+|stage\s+\d+)\s*[:-]?\s*([^.!?]*[.!?])",
            r"(?:successfully|finally)\s+([^.!?]*[.!?])",
        ]

    # Try to find pattern matches
    for pattern in patterns:
        matches = re.findall(pattern, cleaned_content, re.IGNORECASE)
        if matches:
            milestone = matches[0].strip()
            words = milestone.split()[:max_words]
            return " ".join(words) + ("..." if len(milestone.split()) > max_words else "")

    # Fallback: extract first meaningful sentence
    sentences = cleaned_content.split(". ")
    if sentences:
        first_sentence = sentences[0].strip()
        words = first_sentence.split()[:max_words]
        return " ".join(words) + ("..." if len(first_sentence.split()) > max_words else "")

    # Final fallback
    words = cleaned_content.split()[:max_words]
    return " ".join(words) + ("..." if len(cleaned_content.split()) > max_words else "")


def extract_agent_contexts_hierarchical(
    conversation_history: List[Dict[str, Any]], dataset_name: str = ""
) -> List[Dict[str, Any]]:
    """
    Extract hierarchical agent contexts from conversation history.
    Uses graduated detail levels based on distance from current agent.

    Args:
        conversation_history: List of conversation turns with agent information
        dataset_name: Name of the dataset being processed (affects how agent info is extracted)

    Returns:
        List of dictionaries containing hierarchical context for each agent
    """

    contexts = []
    for current_idx in range(len(merged_history)):
        current_agent = conversation_history[current_idx]

        # Build hierarchical context for this agent
        hierarchical_context = {
            "current_agent": current_agent,
            "context_levels": {
                "immediate": [],  # Distance 1: Full detail
                "nearby": [],  # Distance 2-3: Key decisions
                "distant": [],  # Distance 4-6: Brief summaries
                "milestones": [],  # Distance >6: Milestones
            },
        }

        # Process all other agents based on their distance
        for i, agent in enumerate(conversation_history):
            if i == current_idx:
                continue  # Skip current agent

            distance = abs(i - current_idx)

            agent_info = {
                "index": i,
                "name": agent["name"],
                "role": agent["role"],
                "distance": distance,
            }

            if distance == 1:  # Immediate context: Full detail
                agent_info["content"] = agent["content"]
                agent_info["detail_level"] = "full"
                hierarchical_context["context_levels"]["immediate"].append(agent_info)

            elif distance <= 3:  # Nearby context: Key decisions
                agent_info["content"] = extract_key_decision(agent["content"])
                agent_info["detail_level"] = "key_decisions"
                hierarchical_context["context_levels"]["nearby"].append(agent_info)

            elif distance <= 6:  # Distant context: Brief summaries
                agent_info["content"] = summarize_agent(agent["content"])
                agent_info["detail_level"] = "summary"
                hierarchical_context["context_levels"]["distant"].append(agent_info)

            else:  # Milestone context: High-level milestones for very distant agents
                agent_info["content"] = obtain_milestones(agent["content"])
                agent_info["detail_level"] = "milestones"
                hierarchical_context["context_levels"]["milestones"].append(agent_info)

        # Sort each level by original conversation order
        for level in hierarchical_context["context_levels"].values():
            level.sort(key=lambda x: x["index"])

        contexts.append(hierarchical_context)

    return contexts
\end{lstlisting}

\subsection{Context Step Aware Agent}\label{appendix:context_step_aware}
\begin{lstlisting}[language=Python, 
                   basicstyle=\ttfamily\footnotesize,
                   breaklines=true,
                   columns=flexible,
                   backgroundcolor=\color{gray!5}
                   ]

class ContextAwareStepAgent:
    """
    Context-Aware Step Agent that analyzes an agent in the context of its previous and next agents
    to argue why the error happened in this agent's step.
    """

    def __init__(
        self,
        model_id: str = "us.anthropic.claude-3-5-sonnet-20241022-v2:0",
        temperature: float = 1.0,
    ):
        """
        Initialize the Context-Aware Step Agent.

        Args:
            model_id: The model ID to use for the agent
            temperature: The temperature to use for generation
        """
        self.bedrock_model = BedrockModel(
            model_id=model_id,
            temperature=temperature,
            top_p=0.9,
            max_tokens=4096,
        )

        self.system_prompt = """
        You are a Context-Aware Step Agent that analyzes an agent's actions in the context of the previous and next agents.
        Your task is to argue why the error happened in YOUR agent's step.
        
        Your task:
        1. Analyze what information was received by your agent from the previous agent (if any)
        2. Analyze what information was generated by your agent
        3. Analyze how your agent's output affected the next agent (if any)
        4. Make a strong argument for why YOUR AGENT caused the final wrong answer, using the ground truth as evidence
        
        Input:
        - Ground Truth: [GROUND_TRUTH]
        - Final Answer: [FINAL_ANSWER]
        - Agent Context: Information about the previous, current, and next agents
        
        Output your response with the following clear section headers:
        
        ## Purpose:
        Describe the purpose of this agent step - what was this agent trying to accomplish?
        
        ## Assumptions and Information:
        List the assumptions and information this agent was given from the previous agent or context.
        
        ## Errors:
        Describe what this agent did wrong (if anything). Be specific about any mistakes, misunderstandings, or incorrect reasoning.
        
        ## Evidence:
        Provide evidence from the ground truth that supports your error attribution. Explain how this agent's actions directly led to the wrong final answer.
        
        Remember: You must argue that YOUR agent caused the error. Be persuasive and use evidence.
        """

        self.agent = Agent(
            system_prompt=self.system_prompt,
            model=self.bedrock_model,
        )

    def analyze_agent(
        self,
        step_id: str,
        prev_agent: Optional[Dict[str, Any]],
        current_agent: Dict[str, Any],
        next_agent: Optional[Dict[str, Any]],
        ground_truth: Optional[str],
        final_answer: str,
        query: str = "",
    ) -> Dict[str, Any]:
        """
        Analyze an agent in the context of its previous and next agents and generate an argument
        for why this agent caused the error.

        Args:
            step_id: The ID of the step (e.g., "step_1")
            prev_agent: The previous agent (or None if first agent)
            current_agent: The current agent being analyzed
            next_agent: The next agent (or None if last agent)
            ground_truth: The ground truth answer
            final_answer: The final answer given
            query: The original query/question

        Returns:
            JSON argument for why this agent caused the error
        """

        prompt = f"""
        Original Query: {query}
        {ground_truth_section}
        Final Answer: {final_answer}
        Agent Context:
        {agent_context}
        
        Please analyze this agent in the context of the previous and next agents, and provide a strong argument for why THIS agent caused the final wrong answer.
        Use the section headers specified in your instructions (Purpose, Assumptions and Information, Errors, Evidence).
        """

        agent_result = self.agent(prompt)

        # Extract text from AgentResult
        response_text = ""
        if hasattr(agent_result, "message") and "content" in agent_result.message:
            content = agent_result.message["content"]
            if isinstance(content, list) and len(content) > 0 and "text" in content[0]:
                response_text = content[0]["text"]
            elif isinstance(content, str):
                response_text = content

        # Create a dictionary with the step_id, agent_name, and the full text response
        result = {
            "step_id": step_id,
            "agent_name": current_agent["name"],
            "analysis": response_text,
            "token_usage": token_usage,
        }

        return result

    def analyze_agent_hierarchical(
        self,
        step_id: str,
        hierarchical_context: Dict[str, Any],
        ground_truth: Optional[str],
        final_answer: str,
        query: str = "",
    ) -> Dict[str, Any]:
        """
        Analyze an agent using hierarchical context and generate an argument
        for why this agent caused the error.

        Args:
            step_id: The ID of the step (e.g., "step_1")
            hierarchical_context: Dictionary containing hierarchical context information
            ground_truth: The ground truth answer
            final_answer: The final answer given
            query: The original query/question

        Returns:
            JSON argument for why this agent caused the error
        """
        current_agent = hierarchical_context["current_agent"]

        prompt = f"""
        Original Query: {query}
        {ground_truth_section}
        Final Answer: {final_answer}
        Agent Context:
        {agent_context}
        
        Please analyze this agent in the hierarchical context of the entire conversation, and provide a strong argument for why THIS agent caused the final wrong answer.
        Use the section headers specified in your instructions (Purpose, Assumptions and Information, Errors, Evidence).
        
        Note: You now have access to the full conversation context at different detail levels:
        - Immediate context: Full details of adjacent agents
        - Nearby context: Key decisions from agents 2-3 steps away
        - Distant context: Brief summaries of agents 4+ steps away
        
        Consider how information and errors might have propagated across the entire conversation when making your argument.
        """

        agent_result = self.agent(prompt)

        # Extract text from AgentResult
        response_text = ""
        if hasattr(agent_result, "message") and "content" in agent_result.message:
            content = agent_result.message["content"]
            if isinstance(content, list) and len(content) > 0 and "text" in content[0]:
                response_text = content[0]["text"]
            elif isinstance(content, str):
                response_text = content

        # Create a dictionary with the step_id, agent_name, and the full text response
        result = {
            "step_id": step_id,
            "agent_name": current_agent["name"],
            "analysis": response_text,
            "context_type": "hierarchical",
            "token_usage": token_usage,
        }

        return result

\end{lstlisting}

\subsection{Objective Analysis Agent}\label{appendix:objective_analysis}
\begin{lstlisting}[language=Python, 
                   basicstyle=\ttfamily\footnotesize,
                   breaklines=true,
                   columns=flexible,
                   backgroundcolor=\color{gray!5}
                   ]

class ObjectiveAnalysisAgent:
    """
    Objective Analysis Agent that analyzes all agents in a conversation objectively
    to determine error attribution without forced bias.
    """

    def __init__(
        self,
        model_id: str = "us.anthropic.claude-3-5-sonnet-20241022-v2:0",
        temperature: float = 0.7,
        analyst_focus: str = "general",
    ):
        """
        Initialize the Objective Analysis Agent.

        Args:
            model_id: The model ID to use for the agent
            temperature: The temperature to use for generation
        """
        self.bedrock_model = BedrockModel(
            model_id=model_id,
            temperature=temperature,
            top_p=0.9,
            max_tokens=4096,
        )

        # Create specialized system prompt based on analyst focus
        focus_instructions = self._get_focus_instructions(analyst_focus)

        self.system_prompt = f"""
        You are an Objective Analysis Agent conducting an impartial investigation to determine error attribution in a multi-agent conversation.
        
        ANALYST SPECIALIZATION: {focus_instructions}
        
        Your task:
        1. Analyze ALL agents in the conversation objectively (not just one specific agent)
        2. Determine which agent(s) most likely caused the final wrong answer
        3. Determine which step/turn in the conversation the mistake occurred
        4. Provide confidence scores and reasoning for your conclusions
        
        You have access to hierarchical context showing:
        - Immediate agents: Full details
        - Nearby agents: Key decisions
        - Distant agents: Brief summaries
        
        The agents are numbered sequentially (Agent 1, Agent 2, etc.) corresponding to their step/turn index in the conversation.
        
        Possible conclusions:
        - Single agent error: One specific agent caused the mistake at a specific step
        - Multi-agent error: Multiple agents contributed to the mistake across specific steps
        
        Output your response as valid JSON wrapped in <json></json> tags:
        
        <json>
        {{
          "analysis_summary": "Brief overview of your investigation approach and findings",
          "agent_evaluations": [
            {{
              "agent_name": "agent_name",
              "step_index": 1,
              "error_likelihood": 0.0-1.0,
              "reasoning": "Why this agent may or may not have caused the error",
              "evidence": "Specific evidence supporting your assessment"
            }}
          ],
          "primary_conclusion": {{
            "type": "single_agent" | "multi_agent",
            "attribution": ["agent_name(s)"] or null,
            "mistake_step": 1,
            "confidence": 0.0-1.0,
            "reasoning": "Detailed explanation of your primary conclusion including which step the error occurred"
          }},
          "alternative_hypotheses": [
            {{
              "type": "conclusion_type",
              "attribution": ["agent_name(s)"] or null,
              "mistake_step": 1,
              "confidence": 0.0-1.0,
              "reasoning": "Alternative explanation"
            }}
          ]
        }}
        </json>
        
        Be thorough, objective, and consider all possibilities including that no single agent may be clearly at fault.
        Pay special attention to identifying the specific step/turn where the error occurred.
        """

        self.agent = Agent(
            system_prompt=self.system_prompt,
            model=self.bedrock_model,
        )

    def analyze_conversation(
        self,
        conversation_contexts: List[Dict[str, Any]],
        ground_truth: Optional[str],
        final_answer: str,
        query: str = "",
        conversation_history: Optional[List[Dict[str, Any]]] = None,
    ) -> Dict[str, Any]:
        """
        Analyze the entire conversation objectively to determine error attribution.

        Args:
            conversation_contexts: List of hierarchical context dictionaries for all agents
            ground_truth: The ground truth answer
            final_answer: The final answer given
            query: The original query/question

        Returns:
            Dictionary containing objective analysis results
        """

        # Create a comprehensive context summary for analysis
        context_summary = self._create_conversation_summary(conversation_history)

        prompt = f"""
        Original Query: {query}
        {ground_truth_section}
        Final Answer: {final_answer}
        
        Conversation Analysis:
        {context_summary}
        
        Please conduct an objective analysis of this conversation to determine error attribution.
        Focus on identifying which specific agent(s) caused the error that led to the incorrect final answer.

        Output your analysis in the JSON format specified in your instructions.
        """

        agent_result = self.agent(prompt)

        # Extract text from AgentResult
        response_text = ""
        if hasattr(agent_result, "message") and "content" in agent_result.message:
            content = agent_result.message["content"]
            if isinstance(content, list) and len(content) > 0 and "text" in content[0]:
                response_text = content[0]["text"]
            elif isinstance(content, str):
                response_text = content

        try:
            # Parse the JSON response
            analysis_result = validate_json(response_text)
            analysis_result["raw_response"] = response_text
            # Add token usage to the result
            if token_usage:
                analysis_result["token_usage"] = token_usage
            return analysis_result
        except ValueError as e:
            print(f"Error parsing objective analysis response: {e}")
            print(f"Raw response: {response_text}")
            # Return a basic structure if parsing fails
            return {
                "analysis_summary": "Error parsing response",
                "agent_evaluations": [],
                "primary_conclusion": {
                    "type": "single_agent",
                    "attribution": None,
                    "confidence": 0.0,
                    "reasoning": "Failed to parse analysis response",
                },
                "alternative_hypotheses": [],
                "raw_response": response_text,
                "token_usage": token_usage,
            }

    def _create_conversation_summary(
        self, conversation_contexts: List[Dict[str, Any]]) -> str:
        """
        Create a comprehensive summary of the conversation for objective analysis.

        Args:
            conversation_contexts: List of hierarchical context dictionaries

        Returns:
            Formatted conversation summary
        """
        summary = []

        # Extract agent information from contexts with their ORIGINAL step indices
        agents_info = []
        step_indices = list(range(len(conversation_contexts)))

        for i, context in enumerate(conversation_contexts):
            current_agent = context["current_agent"]
            agents_info.append(
                {
                    "step_index": step_indices[i]
                    "name": current_agent["name"],
                    "role": current_agent["role"],
                    "content": current_agent["content"],
                }
            )

        # Create structured summary with clear step indexing
        summary.append("=== CONVERSATION AGENTS ===")
        for agent in agents_info:
            summary.append(f"Step {agent['step_index']} - {agent['name']} ({agent['role']}):")
            summary.append(f"{agent['content']}")
            summary.append("")

        # Add context relationships for the first few agents as examples
        summary.append("=== HIERARCHICAL CONTEXT EXAMPLE ===")
        if conversation_contexts:
            sample_context = format_hierarchical_context(conversation_contexts[0])
            summary.append("Context structure for Agent 1 (showing hierarchical detail levels):")
            summary.append(
                sample_context[:1000] + "..." if len(sample_context) > 1000 else sample_context
            )

        return "\n".join(summary)

    def _get_focus_instructions(self, analyst_focus: str) -> str:
        """
        Get specialized instructions based on analyst focus.

        Args:
            analyst_focus: The type of analyst focus

        Returns:
            Specialized instructions string
        """
        focus_map = {
            "conservative": "You are a conservative analyst with high confidence thresholds. Only attribute errors when you have strong, clear evidence. Prefer single-agent attributions over multi-agent ones. Be cautious about making attributions without definitive proof.",
            "liberal": "You are a liberal analyst more willing to make attributions based on reasonable evidence. Consider multi-agent scenarios and subtle errors that might be overlooked. Be open to making attributions even with moderate confidence.",
            "detail_focused": "You are detail-oriented and focus on specific evidence, exact wording, and fine-grained analysis. Look for subtle inconsistencies, minor logical gaps, and precise factual inaccuracies. Prioritize concrete evidence over general patterns.",
            "pattern_focused": "You are focused on recognizing broader patterns and systemic issues in reasoning chains. Look for recurring themes, logical flow problems, and how errors propagate through the conversation. Consider the overall reasoning structure.",
            "skeptical": "You are highly skeptical and question all assumptions. Look for alternative explanations, consider whether apparent errors might be valid reasoning, and examine if the ground truth itself could be questioned. Challenge conventional attributions.",
            "general": "You are a balanced general analyst with no specific specialization. Approach the analysis with broad perspective, considering all types of evidence equally. Look for the most obvious and impactful mistakes based on objective evaluation.",
        }

        return focus_map.get(analyst_focus, focus_map["general"])
\end{lstlisting}

\subsection{Judge Agent}\label{appendix:judge_agent}
\begin{lstlisting}[language=Python, 
                   basicstyle=\ttfamily\footnotesize,
                   breaklines=true,
                   columns=flexible,
                   backgroundcolor=\color{gray!5}
                   ]

class FinalJudgeAgent:
    """
    Final Judge Agent that weighs competing arguments from multiple Context-Aware Step Agents
    to determine the true error attribution.
    """

    def __init__(
        self,
        model_id: str = "us.anthropic.claude-3-5-sonnet-20241022-v2:0",
        temperature: float = 0.0,
    ):
        """
        Initialize the Final Judge Agent.

        Args:
            model_id: The model ID to use for the agent
            temperature: The temperature to use for generation (lower for more deterministic output)
        """
        self.bedrock_model = BedrockModel(
            model_id=model_id,
            temperature=temperature,
            top_p=0.9,
            max_tokens=4096,
        )

        self.system_prompt = """
        You are a Final Judge Agent that weighs competing arguments from multiple Context-Aware Step Agents to determine the true error attribution.
        
        Your task: Each Context-Aware Step Agent has argued why THEIR agent caused the error. Review all arguments and determine which one is most convincing based on evidence and reasoning.
        
        The arguments from each Context-Aware Step Agent are provided in a structured text format with these sections:
        - Purpose: The purpose of the agent step
        - Assumptions and Information: What the agent was given
        - Errors: What the agent did wrong (if anything)
        - Evidence: Evidence supporting the error attribution
        
        Output your response as a valid JSON object wrapped in <json></json> XML tags. The JSON should have the following structure:
        
        <json>
        {
          "mistake_agent": "agent_name",
          "mistake_step": "step_number",
          "mistake_reason": "explanation of why this agent/step caused the wrong final answer, based on the most convincing argument"
        }
        </json>
        
        IMPORTANT RULES:
        1. Your response MUST be a valid, parsable JSON object wrapped in <json></json> tags. Do not include any text outside these tags.
        2. Focus on the agents that are actively making decisions or providing information.
        
        Be thorough in your analysis. Consider the strength of evidence, the logical connection between the error and the final wrong answer, and the causal relationship.
        Work backwards to see where the logic diverged and the error happened.
        """

        self.agent = Agent(
            system_prompt=self.system_prompt,
            model=self.bedrock_model,
        )

    def judge_arguments(
        self,
        agent_arguments: List[Dict[str, Any]],
        ground_truth: str,
        final_answer: str,
        query: str = "",
    ) -> Dict[str, Any]:
        """
        Judge the competing arguments and determine the true error attribution.

        Args:
            agent_arguments: List of arguments from Context-Aware Step Agents
            ground_truth: The ground truth answer
            final_answer: The final answer given
            query: The original query/question

        Returns:
            Final error attribution as a dictionary
        """
        # Format the agent arguments as a JSON string
        agent_arguments_str = json.dumps(agent_arguments, indent=2)

        prompt = f"""
        Original Query: {query}
        Ground Truth: {ground_truth}
        Final Answer: {final_answer}
        
        All Agent Arguments: {agent_arguments_str}
        
        Please review all the arguments from the Context-Aware Step Agents and determine which one is most convincing.
        Output your decision as a valid JSON object wrapped in <json></json> XML tags as specified in your instructions.
        
        IMPORTANT: Your response MUST be a valid, parsable JSON object wrapped in <json></json> tags. Do not include any text outside these tags.
        """

        agent_result = self.agent(prompt)

        # Extract text from AgentResult
        response_text = ""
        if hasattr(agent_result, "message") and "content" in agent_result.message:
            content = agent_result.message["content"]
            if isinstance(content, list) and len(content) > 0 and "text" in content[0]:
                response_text = content[0]["text"]
            elif isinstance(content, str):
                response_text = content

        try:
            result = validate_json(response_text)
            # Add token usage to the result
            if token_usage:
                result["token_usage"] = token_usage
            return result
        except ValueError as e:
            print(f"Error parsing JSON response: {e}")
            print(f"Raw response: {response_text}")
            # Return a basic structure if parsing fails
            return {
                "mistake_agent": "Unknown",
                "mistake_step": "Unknown",
                "mistake_reason": "Error parsing response",
                "token_usage": token_usage,
            }
                   
\end{lstlisting}

\subsection{Consensus Voting}\label{appendix:consensus_voting}
\begin{lstlisting}[language=Python, 
                   basicstyle=\ttfamily\footnotesize,
                   breaklines=true,
                   columns=flexible,
                   backgroundcolor=\color{gray!5}
                   ]

class ConsensusVotingAgent:
    """
    Consensus Voting Agent that aggregates multiple objective analyses
    to determine final error attribution through voting.
    """

    def __init__(self, min_confidence_threshold: float = 0.3):
        """
        Initialize the Consensus Voting Agent.

        Args:
            min_confidence_threshold: Minimum confidence threshold to consider a conclusion
        """
        self.min_confidence_threshold = min_confidence_threshold

    def aggregate_analyses(
        self,
        objective_analyses: List[Dict[str, Any]],
        ground_truth: str,
        final_answer: str,
        query: str = "",
        conversation_history: Optional[List[Dict[str, Any]]] = None,
    ) -> Dict[str, Any]:
        """
        Aggregate multiple objective analyses through consensus voting.

        Args:
            objective_analyses: List of objective analysis results
            ground_truth: The ground truth answer
            final_answer: The final answer given
            query: The original query/question

        Returns:
            Dictionary containing consensus attribution results
        """
        if not objective_analyses:
            return self._create_empty_result()

        # Extract primary conclusions from all analyses
        primary_conclusions = []
        all_agent_evaluations = defaultdict(list)
        all_alternative_hypotheses = []

        for i, analysis in enumerate(objective_analyses):
            if "primary_conclusion" in analysis:
                conclusion = analysis["primary_conclusion"].copy()
                conclusion["analyst_id"] = i
                primary_conclusions.append(conclusion)

            # Collect agent evaluations
            if "agent_evaluations" in analysis:
                for eval_item in analysis["agent_evaluations"]:
                    agent_name = eval_item.get("agent_name")
                    if agent_name:
                        all_agent_evaluations[agent_name].append(
                            {
                                "error_likelihood": eval_item.get("error_likelihood", 0.0),
                                "reasoning": eval_item.get("reasoning", ""),
                                "evidence": eval_item.get("evidence", ""),
                                "analyst_id": i,
                            }
                        )

            # Collect alternative hypotheses
            if "alternative_hypotheses" in analysis:
                for alt_hyp in analysis["alternative_hypotheses"]:
                    alt_hyp_copy = alt_hyp.copy()
                    alt_hyp_copy["analyst_id"] = i
                    all_alternative_hypotheses.append(alt_hyp_copy)

        # Perform consensus voting
        consensus_result = self._perform_consensus_voting(
            primary_conclusions,
            all_agent_evaluations,
            all_alternative_hypotheses,
            conversation_history,
        )

        # Add metadata
        consensus_result.update(
            {
                "num_analysts": len(objective_analyses),
                "query": query,
                "ground_truth": ground_truth,
                "final_answer": final_answer,
                "voting_method": "weighted_confidence_consensus",
            }
        )

        return consensus_result

    def _perform_consensus_voting(
        self,
        primary_conclusions: List[Dict[str, Any]],
        agent_evaluations: Dict[str, List[Dict[str, Any]]],
        alternative_hypotheses: List[Dict[str, Any]],
        conversation_history: Optional[List[Dict[str, Any]]] = None,
    ) -> Dict[str, Any]:
        """
        Perform consensus voting on the analyses.

        Args:
            primary_conclusions: List of primary conclusions from analysts
            agent_evaluations: Dictionary of agent evaluations by agent name
            alternative_hypotheses: List of alternative hypotheses

        Returns:
            Consensus voting results
        """
        # Vote on conclusion types (single_agent, multi_agent) and collect step predictions
        conclusion_votes = defaultdict(list)

        for conclusion in primary_conclusions:
            conclusion_type = conclusion.get("type", "single_agent")
            confidence = conclusion.get("confidence", 0.0)
            mistake_step = conclusion.get("mistake_step")

            if confidence >= self.min_confidence_threshold:
                conclusion_votes[conclusion_type].append(
                    {
                        "confidence": confidence,
                        "attribution": conclusion.get("attribution"),
                        "mistake_step": mistake_step,
                        "reasoning": conclusion.get("reasoning", ""),
                        "analyst_id": conclusion.get("analyst_id"),
                    }
                )

        # Determine winning conclusion type by weighted confidence
        best_conclusion_type = None
        best_conclusion_info = None
        best_weighted_score = 0.0

        for conclusion_type, votes in conclusion_votes.items():
            # Calculate weighted average confidence
            total_confidence = sum(vote["confidence"] for vote in votes)
            avg_confidence = total_confidence / len(votes) if votes else 0.0
            weighted_score = total_confidence  # Total confidence across all analysts

            if weighted_score > best_weighted_score:
                best_weighted_score = weighted_score
                best_conclusion_type = conclusion_type
                best_conclusion_info = {
                    "votes": votes,
                    "avg_confidence": avg_confidence,
                    "total_confidence": total_confidence,
                    "num_votes": len(votes),
                }

        # For single_agent and multi_agent conclusions, determine which specific agents
        final_attribution = None
        if best_conclusion_type in ["single_agent", "multi_agent"] and best_conclusion_info:
            agent_attribution_votes: defaultdict[str, float] = defaultdict(float)
            for vote in best_conclusion_info["votes"]:
                attribution = vote.get("attribution", [])
                if attribution:
                    for agent_name in attribution:
                        agent_attribution_votes[agent_name] += vote["confidence"]

            # Select agents with highest confidence votes
            if agent_attribution_votes:
                # Sort by confidence and take top agents
                sorted_agents = sorted(
                    agent_attribution_votes.items(), key=lambda x: x[1], reverse=True
                )

                if best_conclusion_type == "single_agent":
                    final_attribution = [sorted_agents[0][0]] if sorted_agents else None
                else:  # multi_agent
                    # Take agents with confidence above threshold
                    final_attribution = [
                        agent
                        for agent, conf in sorted_agents
                        if conf >= self.min_confidence_threshold
                    ]

        # Aggregate agent-level evaluations
        aggregated_agent_evaluations = {}
        for agent_name, evaluations in agent_evaluations.items():
            error_likelihoods = [eval_item["error_likelihood"] for eval_item in evaluations]
            avg_error_likelihood = (
                sum(error_likelihoods) / len(error_likelihoods) if error_likelihoods else 0.0
            )

            aggregated_agent_evaluations[agent_name] = {
                "avg_error_likelihood": avg_error_likelihood,
                "num_evaluations": len(evaluations),
                "evaluations": evaluations,
            }

        # Determine winning step using same methodology as agent attribution
        consensus_mistake_step = None
        step_attribution_votes = {}
        if best_conclusion_type in ["single_agent", "multi_agent"] and best_conclusion_info:
            step_votes_dict: defaultdict[int, float] = defaultdict(float)
            for vote in best_conclusion_info["votes"]:
                mistake_step = vote.get("mistake_step")
                if mistake_step is not None:
                    step_votes_dict[mistake_step] += vote["confidence"]

            if (
                step_votes_dict
                and conversation_history is not None
                and len(conversation_history) > 0
            ):

                # Validate predictions against conversation bounds
                validated_steps = []
                for step, conf in step_votes_dict.items():
                    # Ensure step is integer and within bounds
                    if (
                        isinstance(step, int)
                        and 0 <= step < len(conversation_history)
                    ):
                        validated_steps.append((step, conf))

                if validated_steps:
                    sorted_steps = sorted(validated_steps, key=lambda x: x[1], reverse=True)
                    consensus_mistake_step = sorted_steps[0][0]
                else:
                    consensus_mistake_step = None

                step_attribution_votes = dict(step_votes_dict)
            elif step_votes_dict:
                # No conversation history available, proceed normally
                sorted_steps = sorted(step_votes_dict.items(), key=lambda x: x[1], reverse=True)
                consensus_mistake_step = sorted_steps[0][0] if sorted_steps else None
                step_attribution_votes = dict(step_votes_dict)

        # Handle disagreements
        disagreement_info = self._analyze_disagreements(conclusion_votes)

        return {
            "consensus_conclusion": {
                "type": best_conclusion_type or "single_agent",
                "attribution": final_attribution,
                "mistake_step": consensus_mistake_step,
                "confidence": (
                    best_conclusion_info["avg_confidence"] if best_conclusion_info else 0.0
                ),
                "reasoning": (
                    self._synthesize_reasoning(best_conclusion_info)
                    if best_conclusion_info
                    else "No clear consensus reached"
                ),
            },
            "voting_details": {
                "conclusion_votes": dict(conclusion_votes),
                "step_votes": step_attribution_votes,
                "best_weighted_score": best_weighted_score,
                "disagreement_analysis": disagreement_info,
            },
            "agent_evaluations_summary": aggregated_agent_evaluations,
            "alternative_hypotheses": alternative_hypotheses[:5],  # Keep top 5 alternatives
        }

    def _analyze_disagreements(
        self, conclusion_votes: Dict[str, List[Dict[str, Any]]]
    ) -> Dict[str, Any]:
        """
        Analyze disagreements between analysts.

        Args:
            conclusion_votes: Dictionary of conclusion votes

        Returns:
            Disagreement analysis
        """
        num_conclusion_types = len(conclusion_votes)
        # total_votes = sum(len(votes) for votes in conclusion_votes.values())

        # Check for high disagreement
        high_disagreement = num_conclusion_types > 2 and all(
            len(votes) > 0 for votes in conclusion_votes.values()
        )

        # Calculate confidence spread
        all_confidences: List[float] = []
        for votes in conclusion_votes.values():
            all_confidences.extend(vote["confidence"] for vote in votes)

        confidence_spread = max(all_confidences) - min(all_confidences) if all_confidences else 0.0

        return {
            "high_disagreement": high_disagreement,
            "num_different_conclusions": num_conclusion_types,
            "confidence_spread": confidence_spread,
            "requires_review": high_disagreement or confidence_spread > 0.5,
        }

    def _synthesize_reasoning(self, best_conclusion_info: Dict[str, Any]) -> str:
        """
        Synthesize reasoning from multiple analyst votes.

        Args:
            best_conclusion_info: Information about the best conclusion

        Returns:
            Synthesized reasoning
        """
        if not best_conclusion_info or not best_conclusion_info.get("votes"):
            return "No reasoning available"

        votes = best_conclusion_info["votes"]
        num_votes = len(votes)
        avg_confidence = best_conclusion_info["avg_confidence"]

        # Extract common themes from reasoning
        reasonings = [vote.get("reasoning", "") for vote in votes if vote.get("reasoning")]

        if reasonings:
            # Simple synthesis - could be more sophisticated
            synthesis = f"Consensus reached by {num_votes} analysts (avg confidence: {avg_confidence:.2f}). "
            synthesis += f"Primary reasoning: {reasonings[0][:200]}..."
            if len(reasonings) > 1:
                synthesis += (
                    f" Additional supporting analysis from {len(reasonings)-1} other analysts."
                )
        else:
            synthesis = f"Consensus reached by {num_votes} analysts with average confidence {avg_confidence:.2f}."

        return synthesis

    def _create_empty_result(self) -> Dict[str, Any]:
        """
        Create an empty result when no analyses are provided.

        Returns:
            Empty consensus result
        """
        return {
            "consensus_conclusion": {
                "type": "single_agent",
                "attribution": None,
                "confidence": 0.0,
                "reasoning": "No objective analyses provided",
            },
            "voting_details": {
                "conclusion_votes": {},
                "best_weighted_score": 0.0,
                "disagreement_analysis": {
                    "high_disagreement": False,
                    "num_different_conclusions": 0,
                    "confidence_spread": 0.0,
                    "requires_review": True,
                },
            },
            "agent_evaluations_summary": {},
            "alternative_hypotheses": [],
            "num_analysts": 0,
        }

\end{lstlisting}

\end{document}